\pdfoutput=1

\documentclass[11pt]{article}

\usepackage[]{acl2023}

\usepackage{times}
\usepackage{latexsym}

\usepackage[T1]{fontenc}

\usepackage[utf8]{inputenc}

\usepackage{microtype}

\usepackage{inconsolata}

\usepackage{inconsolata}
\usepackage{graphicx}
\usepackage{subfig}
\usepackage{caption}
\usepackage{threeparttable}
\usepackage{tabularx}
\usepackage{multicol}
\usepackage{multirow}
\usepackage{enumitem}
\usepackage{array}
\usepackage{boxhandler}
\usepackage{url}
\usepackage{mathtools}
\usepackage{amsmath}
\usepackage{adjustbox}
\usepackage{changepage}
\usepackage{makecell}
\usepackage{booktabs}

\title{Exploring Affordance and Situated Meaning in Image Captions: A Multimodal Analysis}

\author{
Pin-Er Chen, Po-Ya Angela Wang, Hsin-Yu Chou, 
Yu-Hsiang Tseng, Shu-Kai Hsieh \\
         Graduate Institute of Linguistics, National Taiwan University \\ 
         cckk2913@gmail.com, differe94nt@gmail.com \\  
         r10142008@ntu.edu.tw, seantyh@gmail.com, shukaihsieh@ntu.edu.tw}

\begin{document}

\maketitle
\begin{abstract}
This paper explores the grounding issue regarding multimodal semantic representation from a computational cognitive-linguistic view. We annotate images from the Flickr30k dataset  with five perceptual properties: \textit{Affordance}, \textit{Perceptual Salience}, \textit{Object Number}, \textit{Gaze Cueing}, and \textit{Ecological Niche Association (ENA)}, and examine their association with textual elements in the image captions. Our findings reveal that images with Gibsonian affordance show a higher frequency of captions containing `holding-verbs' and `container-nouns' compared to images displaying telic affordance. \textit{Perceptual Salience}, \textit{Object Number}, and \textit{ENA} are also associated with the choice of linguistic expressions.
Our study demonstrates that comprehensive understanding of objects or events requires cognitive attention, semantic nuances in language, and integration across multiple modalities. We highlight the vital importance of situated meaning and affordance grounding in natural language understanding, with the potential to advance human-like interpretation in various scenarios.
\end{abstract}

\section{Introduction}

With the rapid advancement of (multimodal) language models, there has been an urgent demand for advanced natural human-machine interactions, as users expect more native-like interactions with AI systems. To attain this sophistication in multimodal communication, the challenge of  \textit{multimodal grounding}, i.e., the pairing of language and other modalities (vision, audio, haptics, etc.), as well as active interaction with the world, has emerged both in the natural language processing (NLP) and computer vision communities. 

Basically, \textit{grounding} refers to associating a word or concept with a perceptual experience in the environment, such as an object or event. Recent tasks such as Visual Grounding (VG) or Natural Language Visual Grounding~\footnote{
\url{https://paperswithcode.com/task/natural-language-visual-grounding}}
have attracted increasing attention, aiming to localize objects/regions in images via natural language expressions~\cite{yang2022improving}. Transformer-based approaches and pretrained vision-and-language (VL) models have greatly succeeded in image and video captioning~\cite{sun2019videobert, radford2021learning, li2023blip}. However, it is worth noting that the term \textit{grounding} carries different meanings in the NLP and cognitive science community. As ~\citet{chandu2021grounding} pointed out, the grounding studies in NLP focus more on the \textit{linking} of text to other modalities. In contrast, the later ones emphasize the \textit{cognitive process} by which the speakers build the common ground to share their mutual information. During this cognitive process, a set of abstract symbols acquire meaning through speakers' perceptions and situated actions based on sensorimotor experiences. The process is similarly proposed and elaborated in cognitive linguistics with the concept of \textit{construal}, representing how individuals mentally interpret a situation or scene~\cite{langacker2008cognitive} and account for the choice of alternative linguistic expressions; i.e., two grammatical possibilities for expressing the same situation are two ways of `construing' that situation~\cite{divjak2020construal}. 
 
We hypothesize the construal of scenes involves common sense knowledge of the presented objects and the visuospatial properties in the images. Therefore, this study systematically examines the grounding issue concerning multimodal semantic representation from a computational cognitive-linguistic view. We operationalize the visuospatial information in the images with five perceptual properties and how they  relate to the construal, which is reflected in the presence of two types of textual elements in the captions. Our research addresses the following questions: (1) How do the five perceptual properties in the images correlate? (2) Does the object \textit{Affordance} in an image relate to the distribution of the two types of textual elements (`holding-verbs' and `container-nouns')\footnote{The two types will be defined in Section \ref{sec:data}.} in its captions? and (3) How do the other perceptual properties associate with the usage of these textual elements in the captions?

The rest of the paper is organized as follows. We first review related works on multimodal cognitive linguistics and the five perceptual properties regarding objects and scenes in Section \ref{sec:related-work}. In Section \ref{sec:method}, we illustrate the dataset and the annotation framework regarding the perceptual properties. Additionally, we conduct exploratory analysis (Section \ref{sec:property-correlation} and \ref{sec:text-element-distribution}) and adopt statistical modeling (Section \ref{sec:statistical-modeling}) on the perceptual properties and the textual elements. Finally, Section \ref{sec:conclusion} concludes the paper.\footnote{The dataset, annotation, and analysis in this study will be publicly available at https://github.com/XXX}

\section{Related work}
\label{sec:related-work}
\subsection{Theoretical framework on multimodal cognitive linguistics}

The fundamental assumptions in Cognitive Linguistics are (i) language is an autonomous, self-contained system; (ii) the linguistic structure is usage-based; and (iii) grammar is inherently symbolic conceptualization \citep{croft2004cognitive, langacker2008cognitive, hart2021can}. These recognize that meaning construction can occur through various semiotic forms of expression within language usage. In other words, Cognitive Linguistics is "particularly well-equipped to unite the natural interest of linguistics in the units that define the language systems with the multimodality of language use" \citep{zima2017multimodality}. 

Recently, Cognitive Linguistics has experienced a multimodal turn, focusing on the interplay between visual perception, linguistic expressions, and the collaborative impact on event conception. \citet{hart2021can} examine the phenomenon of \textit{intersemiotic convergence}, which occurs when language and images share similar forms and create a cohesive relation. They also investigate how linguistic expressions and images converge to shape shared construal in conceptualization by exploring various dimensions\footnote{E.g., schematization, viewpoint, window of attention, and metaphor \citep[see also][]{talmy2000cogsem, forceville2008metaphor, langacker2008cognitive, hart2015viewpoint, hart2021can}.}. Similarly, \citet{divjak2020construal} employ a Visual World Paradigm to study how alternative linguistic constructions (i.e., \textit{location/preposition}, \textit{voice}, and \textit{dative}) modulate the distribution of attention and evoke different conceptualizations. These studies highlight the intricate connection between cognitive mechanisms and linguistic behaviors. 

In terms of linguistic expressions, studies have demonstrated that word meanings are rooted in perception. This connection between language and perception has been extensively explored in Frame semantics \citep{fillmore1976frame} and Generative Lexicon Theory (GLT) \citep{pustejovsky1998generative}, particularly in the context of human-object interaction (HOI) tasks, which serve as a solid foundation for addressing the research questions in our study.

In Frame semantics \citep{fillmore1976frame}, the understanding of objects is based on accumulated experiences, represented as frames. Words are comprehended through the conceptual scenes (frames) they evoke. Building on this framework, \citet{belcavello2020frame} apply fine-grained cognitive semantics in multimodal analysis using FrameNet to investigate how visual objects grounded in the aural modality create frames. Additionally, objects are contextualized to establish their habitat \citep{pustejovsky2013dynamic}. The object's habitat, along with the verb's internal event structure, forms the event simulation. \citet{krishnaswamy2016voxsim} apply these insights to the HOI task, modeling events in a computational virtual environment, and suggest that incorporating affordance learning helps address challenges faced in the robot community. 

\subsection{Perceptual property}
\label{sec:related-work-property}
As previously discussed, the contextualization of objects is proposed to establish their habitat \citep{pustejovsky2013dynamic}, influenced by various affordances in \citet {gibson1977theory}'s theory. \textit{\textbf{Affordance}} refers to the actions enabled by an object for an agent, often termed as "action possibilities" within the surroundings. \citet{henlein2023grounding} distinguish affordance into two categories: \textit{Gibsonian} and \textit{telic}, presenting a model that better detects affordances for novel objects and actions. Gibsonian affordance denotes the "mere interaction with an object," for instance, a \textit{cup} provides Gibsonian affordance for \textit{carrying} or \textit{holding}. On the other hand, telic affordance is related to an object's typical use or purpose in a scene, activating a conventionalized function for the agent \citep{pustejovsky2013dynamic}. For example, in a kitchen, a \textit{cup} naturally affords \textit{telic} actions such as \textit{drinking}, \textit{sipping}, or \textit{pouring}.

In addition to \textit{Affordance}, our study incorporates four other perceptual properties: \textit{Perceptual Salience}, \textit{Object Number}, \textit{Gaze Cueing}, and \textit{Ecological Niche Association (ENA)}.  These properties are crucial for understanding the context of the scene and the affordances offered by the objects. A brief review of each property is provided below: 

\textit{\textbf{Object Number}} provides contextual clues for image interpretation; plural objects may have more than a mere cumulative effect \citep{link1983logical}. The distinction between focused attention and global attention modes suggests different processing for singular and plural objects \citep{treisman2006deployment}.

\textit{\textbf{Gaze Cueing}} is included since attentional connections can be established using visual signals, such as the speaker's gesture or gaze, to emphasize information. \citep{enfield2009anatomy}. In situations with deficient speech, the speaker's gestures gain conversational value for the audience \citep{ozer2023gestures}. This can also apply to image observation, where the gaze of agent(s) depicted in the image guides the attention of the image-viewer(s)\footnote{To avoid confusion, we use the term `agent(s)' to denote the agent(s) portrayed in the image. These agents are the ones interacting with the container-like objects in our analyses. In contrast, the term `viewer(s)' is employed to refer to both (1) people who have observed the images and provided captions in the Flickr30k dataset and (2) our annotators who observe the images and annotate them.}. 

\textit{\textbf{Perceptual Salience}} refers to how much attention a perceived object or event attracts, meaning certain features make an object stand out. In language, we often emphasize a specific part of a scene as the main focus \citep{talmy1983language}. This prominence can be conveyed in two ways: firstly, by plainly specifying emphasized semantic elements; and secondly, by inferring the primacy of an event's participants from its internal semantic structure, even when not directly addressed \citep{langacker2008cognitive}. For instance, both \textit{he drives a car} and \textit{he is driving} emphasize the salience of `car', with the latter omitting the noun phrase.

\textit{\textbf{Ecological Niche Association (ENA)}} introduced in this study refers to the conventionality of an object co-occurring with its environment, denoted as the "ecological niche association." This term captures the mutual dependence and co-adaptation between objects and their surroundings in specific ways. ENA expands the concept of \textit{habitat} as a prerequisite~\citep{pustejovsky2013dynamic} for actions, emphasizing the importance of context in shaping the meaning and function of objects. It highlights the dynamic relationship between objects and their contexts, enriching our understanding of object utilization and interpretation in natural language processing and other applications.

\section{Methodology}
\label{sec:method}

\subsection{Data}
\label{sec:data}
This study focuses on exploring the association between grounding attributes in images and the conceptualized semantic representation in the corresponding captions. To achieve this, we manually select images featuring objects resembling containers from the Flickr30k dataset \citep{young2014flickr}\footnote{The Flickr30k dataset, centering around humans engaging in everyday activities and events, consists of 31,783 images commonly employed for image captioning tasks.}. This selection is guided by our interest in object affordance. Each image from the dataset is paired with five captions, resulting in a total of 733 images and 3665 captions.\footnote{Referring to the comments from the reviewers, the captions, as discussed in \cite{young2014flickr}, are provided by five annotators who lack familiarity with the specific entities and situations depicted in each image. .} 

Regarding the captions, they are lemmatized and POS-tagged via spaCy\footnote{https://spacy.io/}. We define two categories of textual elements found in captions: `holding-verbs' and `container-nouns'. The `holding-verbs' includes motion verbs such as \textit{hold, carry, grasp, grip, lift, grab}, and \textit{take}\footnote{To account for the productive verb \textit{take}, as it also frequently occurs in phrases like \textit{take a bite, take a break, take a sip}, and so on, we have used regular expressions to filter out irrelevant constructions.}. These verbs generally indicate physical contact with hands, while the potentially subsequent actions of the agent may not be explicit.\footnote{The holding-verbs are opposed to verbs like \textit{drink} and \textit{sip}, which imply purposeful actions and presuppose the act of holding a container. For example, when the verbs \textit{drink} or \textit{sip} are used in a caption, they implicitly involve a sequence of actions: \textit{holding} the cup, \textit{lifting} it to the mouth, and \textit{drinking} the liquid.} 
On the other hand, the `container-nouns' consist of specific nouns chosen based on the hyponyms of \textit{container} in the English Wordnet~\footnote{https://wordnet.princeton.edu/}, including \textit{cup, mug, beaker, goblet, chalice, teacup, container, bin, tin, glass, pan, pot}, and \textit{bowl}. 

Regarding the images, we have identified and reviewed five visuospatial properties related to how viewers perceive images (see Section \ref{sec:related-work-property}), as we aim to investigate the associations between properties of groundness in images and conceptualized experssions in captions. The details of the properties will be illustrated with the annotation framework in Section \ref{sec:annot-framework}, and the processing flow of the images and captions is displayed in Appendix \ref{fig:appendix-dataflow}.

\subsection{Annotation Framework}
\label{sec:annot-framework}

The annotation of the images with the five properties, as shown in Table \ref{tab:properties}, includes two types of variables: binary labels (as in \textit{Affordance}, \textit{Object Number}, and \textit{Gaze Cueing}) and a rating scale from 1 to 5 (as in \textit{Perceptual Salience} and \textit{ENA}). For example, images are labeled as "T" in \textit{Affordance} when the depicted actions between the agent and container are purposeful, intentional, and active. On the other hand, images are labeled as "G" when the agent and container lack clear, intentional actions. The labeling of \textit{Gaze Cueing} depends on whether the annotator follows the gaze of the agent towards the container upon first viewing. The \textit{ENA} property is based on the conventional relationship between the container and its surrounding context. For instance, an image of `wine glasses in a restaurant' would receive a higher \textit{ENA} rating than that of `wine glasses in a park'.

\begin{table}
\small
\caption{Perceptual properties for annotation of images involving container-like objects.}
\begin{threeparttable}
\begin{tabularx}{\columnwidth}{@{}p{0.18\columnwidth}p{0.15\columnwidth}X@{}}
\toprule
\textbf{Property} & \textbf{Variable} & \textbf{Description} \\ 
\midrule
Affordance & G / T & Whether the object shows Gibsonian affordance (G) or telic affordance (T).  \\ \midrule
Object Number & S / P & Whether the number of container-like objects is singular or plural.  \\ \midrule
Gaze Cueing & Yes / No & Whether the image-viewer accordingly follows the gaze attention of the agent(s) depicted in the image toward the `container-like object(s)'.  \\ \midrule
Perceptual Salience & 1 (low) - 5 (high) & The degree to which an object or event captures attention, specifically referring to the features that cause an object to be visually distinctive.  \\ \midrule
ENA & 1 (low) - 5 (high)  & The degree of conventional interconnectedness and interdependence between an object and its surrounding environment (scene), describing how they relate and co-adapt to each other in specific ways. \\ \bottomrule
\end{tabularx}
\begin{minipage}{\columnwidth}
\small
\textit{Note.} ENA: Ecological Niche Association.
\end{minipage}
\end{threeparttable}
\label{tab:properties}
\end{table}

\begin{table}[h]
\small
\caption{Inter-annotator agreement rate on the five perceptual properties within each group.}
\begin{threeparttable}
\setlength{\tabcolsep}{6pt} 
\begin{tabularx}{\columnwidth}{@{} *{6}{X} @{}} 
\toprule
\textbf{Property} & \textit{Aff} & \textit{ON} & \textit{GC} & \textit{PS} & \textit{ENA} \\ \midrule 
\textbf{G1} & .89 & .92 & .97 & .89 & .68 \\
\textbf{G2} & .72 & .83 & .85 & .40 & .38 \\ \midrule
\textbf{Avg.} & .80 & .88 & .91 & .65 & .53 \\ \bottomrule
\end{tabularx}
\begin{minipage}{\columnwidth}
\small
\textit{Note.} The properties are shown in abbreviated form: Aff: \textit{Affordance}, ON: \textit{Object Number}, GC: \textit{Gaze Cueing}, and \textit{PS}:  \textit{Perceptual Salience}.
\end{minipage}
\end{threeparttable}
\label{tab:annot-agree}
\end{table}

With a clear understanding of these perceptual properties and trial annotations, four linguists are asked to annotate the 733 images. Two linguists annotated the first half of the dataset (G1), while the other two annotated the second half (G2). After annotation, we have normalized the labels\footnote{\textit{Perceptual Salience} and \textit{ENA} are rated on a 1-5 scale, so we group the original labels into three categories (i.e., < 3, 3, and > 3).} and calculated the inter-annotator agreement rate for each group. Table \ref{tab:annot-agree} presents the statistics of inter-annotator agreement, showing high agreement for \textit{Gaze Cueing}, \textit{Object Number}, and \textit{Affordance}. This indicates that the annotations are consistent and suitable for subsequent analysis. In cases where there were disagreements within a group (e.g., G1), we involve annotators from the other group (G2), who have not seen the images, to discuss and provide a third annotation to determine the final labeling for those images.

\section{Discussion}
\subsection{Correlation between Perceptual Properties}
\label{sec:property-correlation}

\begin{figure}[]
\caption{Correlation matrix between the independent properties.}
\includegraphics[width=0.9\columnwidth]{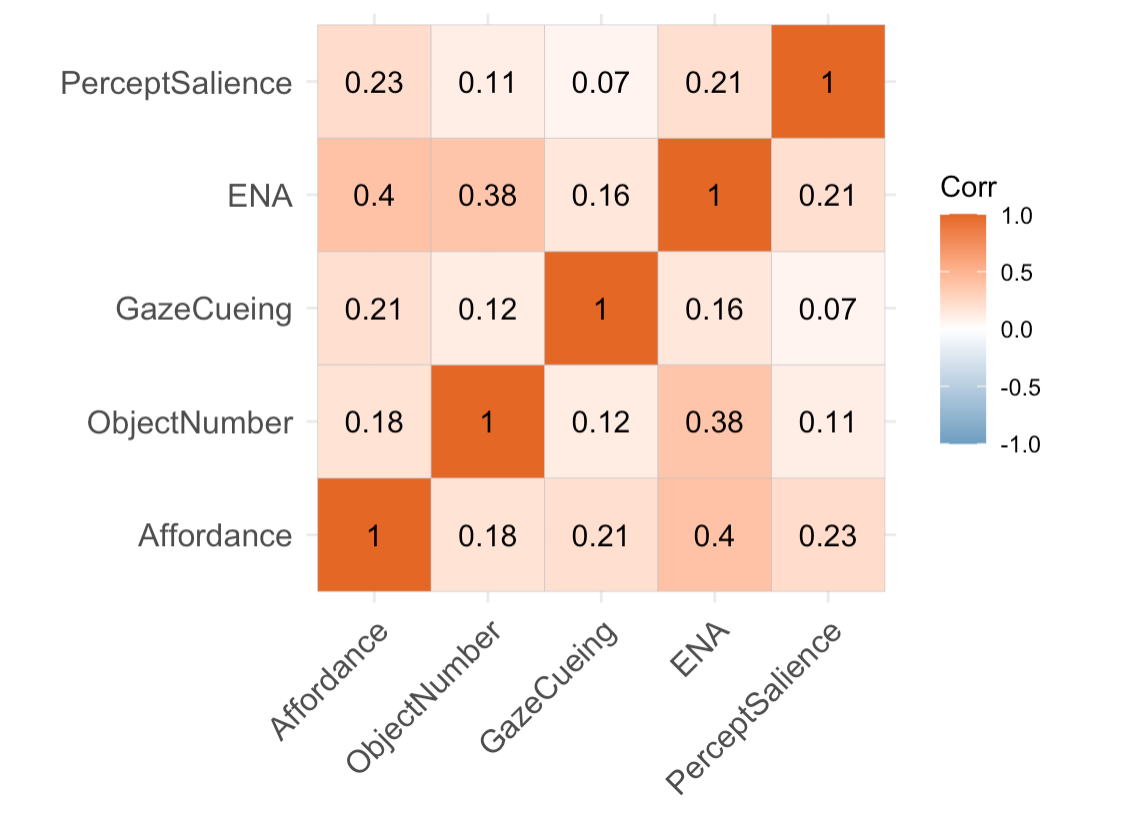}
\label{fig:corr-matrix}
\end{figure}

Firstly, we investigate the correlation structures between the perceptual properties. In 
Figure \ref{fig:corr-matrix}, we compute a matrix of Pearson’s bivariate correlations for each pair of independent properties.~\footnote{In computing the correlation, the values of \textit{Affordance} (i.e., "T" and "G") and \textit{Object Number} (i.e., "P" (plural) and "S" (singular)) are both transformed to 1 and 0.} We follow \citet{mason1991collinearity}'s suggestion in using bivariate correlation 0.8-0.9 as the cutoff threshold, above which indicates strong linear associations and a linearity problem. With all the values being less than 0.8, we are more confident the multi-collinearity will not be a significant issue when interpreting each model predictor's contribution.

Some observations in the correlation matrix are noteworthy. Figure \ref{fig:corr-matrix} shows positive correlations between \textit{ENA} and \textit{Affordance} (corr = 0.4) as well as \textit{ENA} and \textit{Object Number} (corr = 0.38). The positive correlation between \textit{ENA} and \textit{Affordance} indicates that as the container(s) appear more conventional in their natural environment (higher \textit{ENA}), they are more likely to be linked with a telic affordance. This discovery aligns with the concept of \textit{telic} affordance by \citet{henlein2023grounding}, denoted as "a conventionalized configuration to activate a conventionalized function." Namely, a container with purposeful functions in a conventionalized scene is likely to be intentionally used.

A positive correlation between \textit{ENA} and \textit{Object Number} is also observed. In settings such as kitchens or grocery stores, multiple containers are considered a collective noun, which serves as a situational and conventional cue. They become part of the "conventionalized configuration" that activates a customary function \citep{henlein2023grounding}. For instance, in Figure \ref{fig:corr-on-ena} (see Appendix \ref{sec:appendix-ex-pictures}), the numerous containers create the arranged setup, as "the products of the vendor," which prompts the agent to engage in "selling." This configuration of container(s) and functional arrangements leads to a higher \textit{ENA} score.

\subsection{Affordance \& Distribution of Textual Elements}
\label{sec:text-element-distribution}

We delve into the distribution of `holding-verbs' and `container-nouns' concerning the research question: "Is the object \textit{Affordance} in an image related to the presence of the two types of textual elements in its captions?" Our hypothesis proposes that object \textit{Affordance} is associated with the textual elements in linguistic expressions, and that viewers are more likely to use holding-verbs/container-nouns when describing images involving objects with `Gibsonian affordance'. As objects with `telic affordance' imply more purposeful uses, they lead viewers to use more specific and informative language in conceptualization of such images. 

For each image, we quantify the number of captions with holding-verbs and container-nouns separately. As we aim to explore the different \textit{Affordance} of the container-like objects, we categorize the 733 images into two groups: T-group (346) and G-group (387) based on our annotations. "T" denotes the telic affordance, and "G" denotes the Gibsonian affordance.

\begin{table}[!ht]
\small
\caption{Distribution of textual elements for T-group and G-group images.}
\begin{tabular}{p{2.4cm}|p{1cm}||p{1.2cm}|p{1.2cm}} 
\toprule
\multirow{2}{*}{\textbf{Textual Element}} &
  \multirow{2}{*}{\textbf{Cap$_N$}\textsuperscript{*}} &
  \multicolumn{2}{c}{\textbf{Image$_N$ (\%) }\textsuperscript{**}} \\ \cline{3-4} 
  &
  &
  \multicolumn{1}{c}{\textit{T-group}} &
  \multicolumn{1}{c}{\textit{G-group}} 
  \\ \midrule
Holding-verbs &
\begin{tabular}{c} 0/5 \\ 1/5 \\ 2/5 \\ 3/5 \\ 4/5 \\ 5/5\\
\end{tabular} &
\begin{tabular}{c} 67.6 \\ 21.1 \\ 6.1 \\ 2.9 \\ 2.0 \\ 0.3 \\
\end{tabular} &
\begin{tabular}{c} 47.0 \\ 26.4 \\ 13.4 \\ 10.6 \\ 2.1 \\ 0.5 \\
\end{tabular}
\\ \midrule
Container-nouns &
\begin{tabular}{c} 0/5 \\ 1/5 \\ 2/5 \\ 3/5 \\ 4/5 \\ 5/5\\
\end{tabular} &
\begin{tabular}{c} 86.1 \\ 9.8 \\ 2.6 \\ 0.6 \\ 0.3 \\ 0.6 \\
\end{tabular} &
\begin{tabular}{c} 78.3 \\ 15.8 \\ 4.1 \\ 1.6 \\ 0.3 \\ 0.0 \\
\end{tabular} 
\\ \bottomrule
\end{tabular}
\begin{minipage}{\columnwidth}
\small
\textit{Note.} \textsuperscript{*} Cap$_N$: The number of captions containing the target textual elements, out of the five captions per image. 
\textsuperscript{**} Image$_N$: The percentage of number of images; for example, 0.3\% of T-group images and 0.5\% of G-group images contain holding-verbs in all of their captions (5 out of 5 captions per image).
\end{minipage}
\label{tab:element-distribution}
\end{table}

Table \ref{tab:element-distribution} shows interesting patterns in holding-verbs and container-nouns within T-group and G-group image captions. For T-group images, 67.6\% (234) of them do not contain any holding-verbs in captions (i.e., 0 out of 5 captions per image), whereas a significant proportion (86.1\%, 298) of them include no container-nouns in captions. Contrarily, approximately 47\% (182) of G-group image lack holding-verbs in captions, and 78\% (303) of them do not contain any container-nouns in captions. In general, G-group exhibits a higher proportion of images incorporating either holding-verbs or container-nouns in their captions.\footnote{Among the G-group images, 53\% include holding-verbs in at least one caption, while the T-group exhibit a percentage of 32\%. Also, 22\% of G-group images contain container-nouns in at least one caption, while the T-group exhibit a lower percentage of 14\%.} This finding supports our hypothesis of a stronger association between objects with Gibsonian affordance in images and the occurrence of the two textual elements, in comparison to objects with telic affordance.

Besides \textit{Affordance}, we believe that the other aforementioned perceptual properties, regarding human attention and dynamic relationship between container(s) and scene in images, also play essential roles in conceptualized linguistic expressions (see Section \ref{sec:related-work-property}). We attempt to adopt multiple linear regression models to evaluate the association between different perceptual properties and the usage of textual elements in image captions.


\subsection{Statistical Modeling}
\label{sec:statistical-modeling}
We have employed two multiple linear regression models to investigate the question on "how the perceptual properties in an image contribute to the occurrence of textual elements within the captions." Both models include five independent variables (i.e., perceptual properties): \textit{Affordance} (T/G), \textit{Object Number} (Singular/Plural), \textit{Gaze Cueing} (Yes/No), \textit{Perceptual Salience} (1-5), and \textit{ENA} (1-5). The dependent variables are the numbers of captions with holding-verbs or container-nouns for each image, which are normalized to a range of 0 to 1. For example, if an image owns 3 out of 5 captions containing holding-verbs, the value for its holding-verb usage would be 0.6; if the image owns 2 out of 5 captions containing container-nouns, the value for its container-noun usage would be 0.4. The results of the models for holding-verbs and container-nouns will be discussed separately in Section \ref{sec:hold-verbs-model} and \ref{sec:c-nouns-model}, with an overview at the end of Section \ref{sec:c-nouns-model}.

\begin{table}[!h]
\centering
\caption{Results of multiple linear regression model for holding-verbs.}
\begin{threeparttable}
\begin{adjustbox}{width=\columnwidth}
\begin{tabular}{lrrrr}
\toprule
{\textbf{Variable}} & \textbf{Coeff} & \textit{\textbf{SE}} & \textit{\textbf{t}} & \textit{\textbf{P}} \\
\midrule
(Intercept) & -0.01 & 0.035 & -0.296 & .768 \\
\textbf{Perceptual Salience} & 0.063 & 0.007 & 8.886 & <.001 *** \\
\textbf{Object Number\_S} & 0.057 & 0.015 & 3.733 & <.001 *** \\
\textbf{ENA} & -0.021 & 0.007 & -2.881 & <.005 ** \\
\textbf{Affordance\_T} & -0.073 & 0.016 & -4.608 & <.001 *** \\
\textbf{Gaze Cueing} & -0.096 & 0.018 & -5.414 & <.001 *** \\
\bottomrule
\end{tabular}
\end{adjustbox}
\begin{minipage}{\columnwidth}
\small
\textit{Note.} Affordance\_T: \textit{Affordance} labeled as T (telic). Object Number\_S: \textit{Object Number} labeled as S (singular).
\end{minipage}
\end{threeparttable}
\label{tab:stat-verb}
\end{table}

\renewcommand{\arraystretch}{1.5}
\begin{table*}[!h]
  \centering
  \caption{Example images with captions: (A) above and (B) below}
  \begin{tabularx}{\textwidth}{m{4.5cm} m{20cm} }
    \toprule
    \textbf{Example Image (A) \& (B)} & \textbf{Captions} \\ 
    \midrule
    \begin{minipage}{\columnwidth}
      \includegraphics[width=45mm, height=30mm]{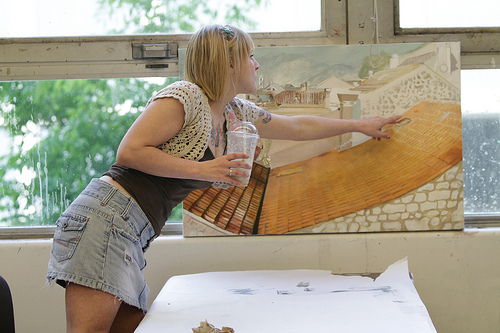}
    \end{minipage}
    &
    \begin{minipage}{24cm}
     \small
   \begin{enumerate}[rightmargin=13cm]
        \item A blond woman in a short denim skirt, black top, and beige jacket, is reaching toward a part of a painting that is propped up on a windowsill.
        \item A woman is \textbf{holding a drink} in one hand and pointing at a painting with the other.
        \item Woman with a jean skirt \textbf{holding a drink} points to an object in a painting.
        \item A girl \textbf{holding an empty plastic cup} is pointing to a painting.
        \item A girl \textbf{holding a beverage} points at a painting.
      \end{enumerate}
    \end{minipage}
    \\ 
    \midrule
    \begin{minipage}{\columnwidth}
      \includegraphics[width=45mm, height=30mm]{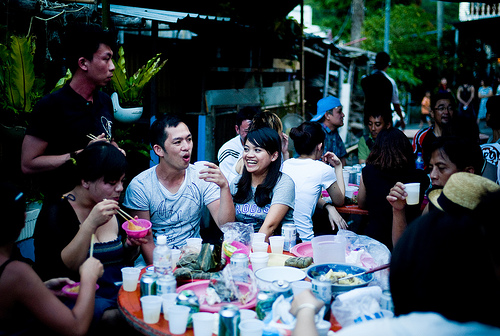}
    \end{minipage}
    &
    \begin{minipage}{24cm}
     \small
   \begin{enumerate}[rightmargin=13cm]
        \item A man and a smiling woman sit at a dining table with many plastic \textbf{cups} on it as a person next to them eats out of a \textbf{bowl} with chopsticks.
        \item A group of people eat a meal in a crowded outdoor location.
        \item A group of people enjoy food and drinks at an outdoor party.
        \item A group of people eating and talking around a table.
        \item People are gathered at a table to enjoy drinks.
      \end{enumerate}
    \end{minipage}
    \\ 
    \bottomrule
  \end{tabularx}
  \label{tab:example-images}
\end{table*}

\subsubsection{Regression model for holding-verbs}
\label{sec:hold-verbs-model}

The results of the model are presented in Table \ref{tab:stat-verb}.  Firstly, \textit{Perceptual Salience} shows a positive relationship with the holding-verbs usage (estimate: 0.0625, p < .001). This suggests that when a container in an image is attention-grabbing to viewers, there is a higher likelihood for viewers to use holding-verbs in  captions. For instance, a cup is at the center of image (A) in Table \ref{tab:example-images}. While the cup is apparently not the female’s attention nor the main theme of the image, it tends to be mentioned in the captions. Among the five captions for (A), as shown in the upper row of Table \ref{tab:example-images}, four of them contain the verb \textit{hold}; specifically, the majority of the captions involve participle constructions (i.e., \textit{holding a drink; holding a cup; holding a beverage}) to modify the agent (i.e., \textit{the woman; the girl}).

Regarding \textit{Object Number}, a significant positive correlation with holding-verbs usage is observed (estimate: 0.0574, p = .0002). This suggests a strong tendency for viewers to use holding-verbs in captions when a singular container appears in an image. In contrast, \textit{ENA} shows a negative relationship with holding-verbs usage (estimate: -0.0208, p = .0041). This implies that holding-verbs are more likely to be employed by viewers when an image portrays a scene that is less conventional. For instance, Figure \ref{fig:ena-1} (see Appendix \ref{sec:appendix-ex-pictures}) presents a scene where the conventional function of the cup is not evident (i.e., low \textit{ENA}). In this case, three out of five captions contain the act of \textit{carrying a drink}, \textit{holding a cup}, and \textit{holding a coffee cup}, while the other two captions do not refer to the cup. On the other hand, Figure \ref{fig:ena-5} (see Appendix \ref{sec:appendix-ex-pictures}) displays a scenario where multiple containers are situated within a kitchen or party setting. It reflects higher conventionality of the containers co-occurring with their environment (i.e., high \textit{ENA}), which aligns with \citet{pustejovsky2013dynamic}'s definition of a habitat as the precondition for an action involving the object. This activation prompts viewers to use verbs that directly describe the container's function, such as \textit{drink, sip, pour, stir}, rather than holding-verbs.

In terms of \textit{Affordance}, the coefficient for the \textit{Affordance\_T} variable is negatively significant (p < .001). This is because annotators assign the "T" label for \textit{Affordance} when the depicted relations between the agent and container in images seem telic and purposeful. As these actions are explicit, it is reasonable for viewers to choose more specific verbs rather than less specific holding-verbs in their captions. Conversely, images are labeled as "G" when annotators perceive no clear intentional actions. As agent(s) of these images typically has mere contact with the container, i.e., "behaviors afforded due to the physical object structure" \cite{henlein2023grounding}, viewers tend to use more general verbs, the holding-verbs, to describe the relationship between the agent and the container.

As for \textit{Gaze Cueing}, it shows a significant negative relationship with the use of holding-verbs (estimate: -0.0956, p < .001). This suggests that when the agent of an image employs explicit \textit{Gaze Cueing}, directing viewer's attention, viewers are less likely to use holding-verbs in captions. In Figure \ref{fig:gaze-y} (see Appendix \ref{sec:appendix-ex-pictures}), the agents look directly at the containers (i.e., \textit{Gaze Cueing}: yes), showing intentional engagement with the containers; the captions for this image include telic verbs like \textit{mix}, \textit{pour}, and \textit{perform}, rather than holding-verbs. Conversely, in Figure \ref{fig:gaze-n} (see Appendix \ref{sec:appendix-ex-pictures}), the agents' gaze is not at the container but at the screen (i.e., \textit{Gaze Cueing}: no)\footnote{It is noted that \textit{Gaze Cueing} in this study only represents the agent's gaze attention toward "container-like object(s)." The agent's gaze at other objects may be taken into account as another type of \textit{Gaze Cueing} in future studies.}. In this case, viewers tend to use verbs related to `looking' as the main action and use holding-verbs only to modify the agent (in relation to the container).


\subsubsection{Regression model for container-nouns}
\label{sec:c-nouns-model}
We also conducted multiple linear regression analysis to examine the usage of container-nouns. The same variables as in the model for holding-verbs were utilized, as displayed in Table \ref{tab:stat-noun}.

\begin{table}[!h]
\centering
\small
\caption{Results of multiple linear regression model for container-nouns.}
\begin{threeparttable}
\begin{adjustbox}{width=\columnwidth}
\begin{tabular}{lrrrr}
\toprule
{\textbf{Variable}} & \textbf{Coeff} & \textit{\textbf{SE}} & \textit{\textbf{t}}& \textit{\textbf{P}} \\ 
\hline
(Intercept) & -0.055  &  0.114 & -0.482 & .63\\
\textbf{Perceptual Salience} & 0.107 & 0.023 &  4.701 & <.001 *** \\
\textbf{Object Number\_S} & 0.182 &  0.05 & 3.649 & <.001 *** \\
\textbf{ENA}  &  -0.061 &  0.023 & -2.632 & .008 **    \\
\textbf{Affordance\_T} & -0.047  &  0.052 & -0.921 & .358\\
\textbf{Gaze Cueing}   & 0.021 &  0.057 & -0.374  & .708 \\
\bottomrule            
\end{tabular}
\end{adjustbox}
\begin{minipage}{\columnwidth}
\small
\textit{Note.} Affordance\_T: \textit{Affordance} labeled as T (telic). Object Number\_S: \textit{Object Number} labeled as S (singular).
\end{minipage}
\end{threeparttable}
\label{tab:stat-noun}
\end{table}

For each unit increase in \textit{Perceptual Salience}, there is a positive estimate of 0.1073 (p < .001) in the number of captions containing container-nouns. This suggests that when a container in an image is more visually noticeable, viewers tend to use container-nouns more frequently in their captions. This aligns with our earlier discussion on holding-verbs. Even though the container is not the primary focus of the scene, its salience prompts viewers to include its description when conceptualizing the image. Consequently, container-nouns (e.g., \textit{cup}) are used in participial phrases to modify the main agent/focus of the image, as in caption 4 (\textit{A girl `holding an empty plastic cup' is pointing to a painting.}) in the upper row of Table \ref{tab:example-images}.

The \textit{Object Numbe\_S} (singular) also demonstrates a significant positive relationship with the use of container-nouns (estimate: 0.1821, p < .001), suggesting that when a solitary container is presented, viewers tend to use container-nouns more frequently in captions. This preference arises from the ability to concentrate attention on a singular object, leading to the expectation of more precise distinctions \cite{treisman2006deployment}. In contrast, if the number of container in an image is plural, the captions are less likely to include container-nouns. This can be observed in image (B) in Table \ref{tab:example-images}. In scenarios with an abundance of container, such as in a café or gathering, the individual significance and distinctiveness of containers decrease. Viewers tend to either concentrate on describing specific elements of the scene (e.g., agent(s) engaged in a purposeful action) or depict the scene as a whole. This can be seen in captions 2-5 in the second row of Table \ref{tab:example-images}.

On the contrary, \textit{ENA} shows a slightly significant negative relationship, with an estimate of -0.0616 (p = .008). This indicates that when an image depicts a less conventional scene, viewers tend to use container-nouns more frequently in captions. This observation is consistent with findings concerning \textit{Object Number}. In Figure \ref{fig:ena-1} (see Appendix \ref{sec:appendix-ex-pictures}), where there is only one cup and a scene difficult for viewers to identify the conventional function (i.e., singular object \& low \textit{ENA}), the captions contain more phrases with container-nouns (e.g., \textit{holding a cup}). In contrast, captions for Figure \ref{fig:ena-5} contain fewer container-nouns as this image presents an accumulation of containers and a scene with higher conventionality that can be easliy identified as a party (i.e., plural objects \& high \textit{ENA}). As for the other variables, \textit{Affordance (T)} and \textit{Gaze Cueing} did not exhibit statistical significance.

\begin{table}[!h]
\centering
\small
\begin{threeparttable}
\caption{Statistically significant factors for the presence of holding-verbs and container-nouns in captions.}
\begin{tabularx}{\columnwidth}{lX}
\toprule
\textbf{Holding-verbs} & \\
\hline
\textit{Perceptual Salience} & The container is perceptually noticeable to viewer (high). \\
\hline
\textit{Object Number} & The number of the container is singular (S). \\
\hline
\textit{Gaze Cueing} & The agent does not employ explicit gaze cueing to the container (low). \\
\hline
\textit{ENA} & The scene depicted in the image is less conventional (low). \\
\hline
\textit{Affordance}& The object shows Gibsonian affordance (G). \\
\hline\hline
\textbf{Container-nouns} & \\
\hline
\textit{Perceptual Salience} & The container is perceptually noticeable (high). \\
\hline
\textit{Object Number} & The number of the container is singular (S). \\
\hline
\textit{ENA} & The scene depicted in the image is less conventional (low). \\
\bottomrule
\end{tabularx}
\label{tab:model-result-summary}
\end{threeparttable}
\end{table}

Table \ref{tab:model-result-summary} presents a summary of significant factors in the two models, highlighting specific properties in images that prompt viewers to use these textual elements more frequently in captions. The results strongly support our hypothesis, indicating a preference for holding-verbs in conceptualizing objects with Gibsonian affordance. When viewers observe an image depicting agent(s) and container(s), they determine if the container serves a purposeful function for the agent in such scene. If it does not, i.e., indicating  Gibsonian affordance, viewers tend to use holding-verbs like \textit{hold} or \textit{take} to describe the container while modifying the agent (e.g., girl \textit{holding a glass}). In terms of the other perceptual properties, \textit{Perceptual Salience} and \textit{Object Number\_S} exhibit significantly positive relationships with the usage of the two textual elements, while \textit{ENA} shows less significant negative correlation with them; \textit{Gaze Cueing} shows significant negative relationship only with the usage of holding-verbs. They facilitate the dynamic convergence between the container and its habitat \citep{pustejovsky2013dynamic} within the image, improve context comprehension, and contribute to the selection of linguistic expression. Overall, the analyses highlight the crucial role played by human cognitive mechanisms, object affordance, and contextual information in shaping shared construal by integrating visually-perceived events and text \citep{hart2021can}.



\section{Conclusion}
\label{sec:conclusion}
This study investigates the grounding issue in multimodal semantic representation, focusing on five perceptual properties in images and their associations with two types of textual elements in captions. Regarding \textit{Affordance}, images featuring Gibsonian affordance show higher frequency of captions containing `holding-verbs' and `container-nouns' compared to images featuring telic affordance. The other properties, namely \textit{Perceptual Salience}, \textit{ENA}, \textit{Gaze Cueing}, and \textit{Object Number}, also play vital roles in shaping linguistic expressions of scenes. 
Our findings highlight the significance of situated meaning and object affordance in human conceptualization of visual input, transcending mere combination of text and other modalities. They offer insights for computational cognitive science, multimodal communication, and the contextually grounded AI models.

Despite limitations such as subjective selection of target images and the need for evaluations of provided captions and annotations, our study open up possibilities for bidirectional tasks involving visual and textual elements for machines. Regarding future work, we plan to extend our research to multimodal datasets that contain scenes where the visual cues and affordance are not as obvious or entirely absent, ensuring that the insights 
we've gained can be applied beyond images with clear affordance. 
To effectively handle situations where images lack evident affordance, we will explore the incorporation of additional contextual cues and the advanced deep learning techniques, which will help us bridge the gap between the visual characteristics of scenes and the language used to describe them in more intricate visual contexts. Overall, by integrating situatedness into multimodal semantics, we can improve our understanding of human interpretation in diverse real-world situations and facilitate further research on groundedness in natural language understanding systems.

\bibliography{anthology,custom}
\bibliographystyle{acl_natbib}

\appendix
\begin{minipage}{0.9\columnwidth}
    \centering
    \section{Data Processing Flow}
        \includegraphics[width=\columnwidth, height=0.8\textheight, keepaspectratio]{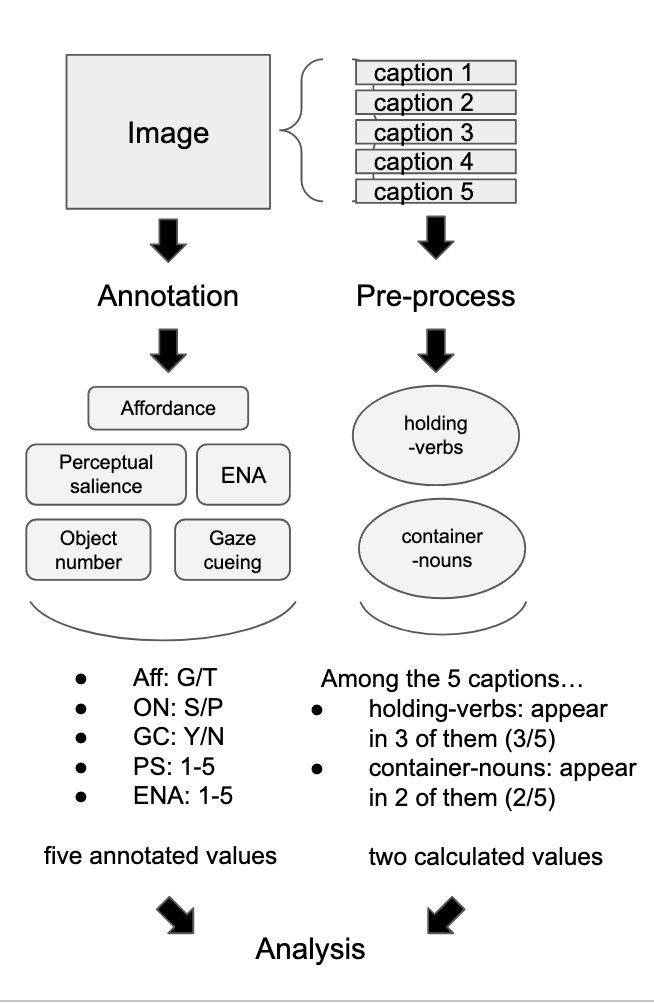}
        \label{fig:appendix-dataflow}
\end{minipage}

\section{Example Pictures}
 \label{sec:appendix-ex-pictures}
\begin{figure}
  \centering
  \subfloat[Correlation between Object Number and ENA.]{\includegraphics[width=0.49\linewidth]{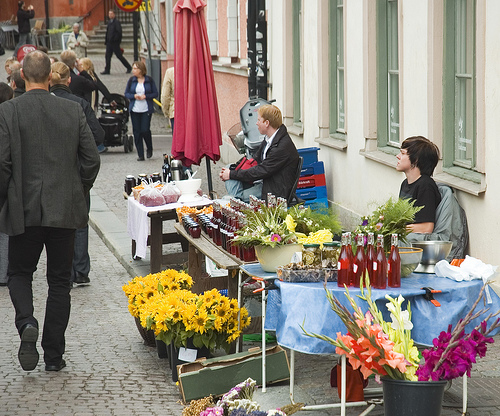}\label{fig:corr-on-ena}}
  \hfill
  \subfloat[ENA: 1]{\includegraphics[width=0.49\linewidth,height=50mm]{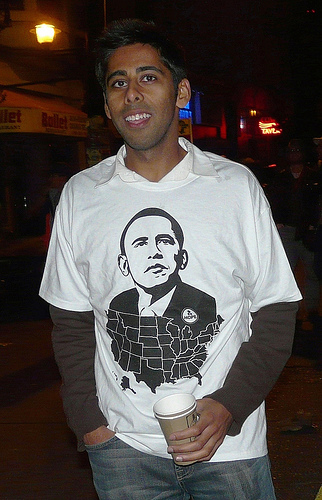}\label{fig:ena-1}}
  \hfill
  \subfloat[ENA: 5]{\includegraphics[width=0.49\linewidth]{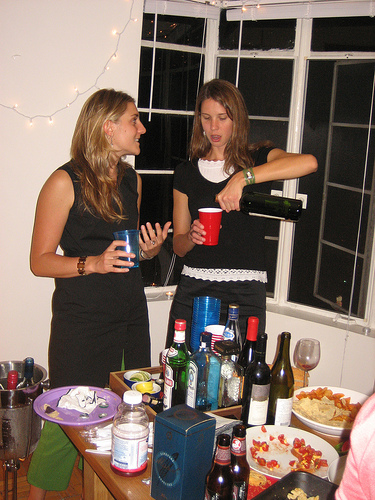}\label{fig:ena-5}}
  \hfill
  \subfloat[Gaze Cueing: Y]{\includegraphics[width=0.49\linewidth]{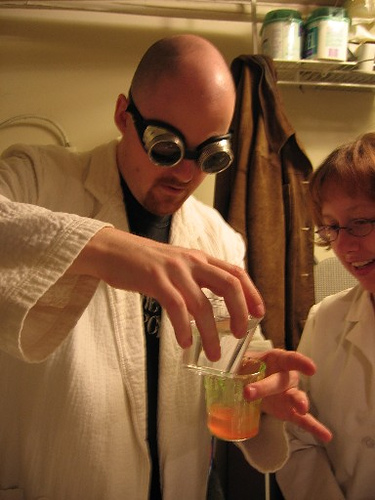}\label{fig:gaze-y}}
  \hfill
  \subfloat[Gaze Cueing: N]{\includegraphics[width=0.49\linewidth]{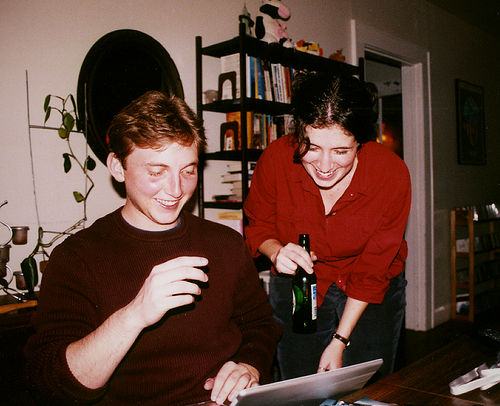}\label{fig:gaze-n}}
  \hfill
  \caption{Example images. The value of \textit{ENA} scales from 1 to 5; The value of \textit{Gaze Cueing} is either Y (Yes) or N (No).}
\end{figure}

\end{document}